# A description length approach to determining the number of k-means clusters

Hiromitsu Mizutani and Ryota Kanai

Araya Inc.

mizutani@araya.org, kanai@araya.org

Abstract

We present an asymptotic criterion to determine the optimal number of clusters in k-means. We consider k-means as data compression, and propose to adopt the number of clusters that minimizes the estimated description length after compression. Here we report two types of compression ratio based on two ways to quantify the description length of data after compression. This approach further offers a way to evaluate whether clusters obtained with k-means have a hierarchical structure by examining whether multi-stage compression can further reduce the description length. We applied our criteria to determine the number of clusters to synthetic data and empirical neuroimaging data to observe the behavior of the criteria across different types of data set and suitability of the two types of criteria for different datasets. We found that our method can offer reasonable clustering results that are useful for dimension reduction. While our numerical results revealed dependency of our criteria on the various aspects of dataset such as the dimensionality, the description length approach proposed here provides a useful guidance to determine the number of clusters in a principled manner when underlying properties of the data are unknown and only inferred from observation of data.

1. Introduction

K-means clustering is widely used for its simplicity and computational efficiency. However, one of the major practical problems of k-means is that the algorithm itself cannot determine the optimal number of clusters. There have been a few proposals as to possible approaches for this problem. For example, regularized k-means (Sun et al., 2012) proposed a regularization term for the k-means algorithm, and modified the algorithm to find a consistent number of clusters. The X-means algorithms (Pelleg and Moore, 2000) is also known as an algorithm to stop the iteration of the clustering using Bayesian Information Criterion (BIC) as a criterion. The results and algorithms of these methods are accurate when the probabilistic distribution of the dataset can be assumed to be some variation of a Gaussian. However, whether this assumption holds for empirical data is generally unclear.

To offer a practical criterion to determine the number of clustering in empirical studies where the nature of the distribution is unknown, we propose the use of compression ratio as a simple criterion to determine the optimal number of clusters without changing the algorithm of k-means. Here, we consider the k-means algorithm as data compression, and evaluate the compression ratio using description length of the dataset, following the idea of minimum description length (MDL). The idea is that the optimal number of clusters should provide the most efficient compression ratio. In this paper, we discuss two types of description length, each of which gives rise to a definition of compression ratio.

Furthermore, the approach to compute description length also offers a systematic way to examine whether clustering was left redundancy in data. That is to say, if further compression is possible after the first-level k-means clustering, the data size after the second compression should be even smaller. However, such cases indicate that redundancy is present in the first-stage k-means clustering, which is violates the assumption of k-means that there is no hierarchical structure in the dataset. Thus, the description length approach also offers a way to reveal potential hierarchical structure in the dataset.

To observe the behavior of the compression ratio and demonstrate the usability of our criteria in practice, we tested the criteria with synthetic and empirical brain MRI data. For the MRI data, we applied the proposed criteria to a brain parcellation problem using structural MRI data. Previously, Glasser et al. (2016) showed a parcellation of human cerebral cortex using both the structural MRI and functional MRI data. Here, we focus on the structural MRI dataset to determine the number of clusters (i.e. brain regions) based on interindividual differences in the local volumetric measure of gray matter (Kanai & Rees, 2011). Since the main focus of the current paper is to present the criteria to determine the number of clusters in k-means, we will not discuss neurobiological implications of the results in details.

## 2. Methods

### 2.1. Notations

Throughout this paper, we type vectors in bold ($\mathbf{x}$), scalars in regular ($R, b$), and matrices in capital bold ($\mathbf{X}$). Specific entries in vectors or matrices are scalars and follow the corresponding convention, i.e. the i-th dimension of vector $\mathbf{x}$ is $x_i$. In contrast, depending on the context, $\mathbf{x}^{(i)}$ refers to the i-th column in matrix $\mathbf{X}$, and $\mathbf{x}_{(i)}$ refers to i-th row.

### 2.2. Description length of k-means

Here, we present two representations (DL1 and DL2) of criteria for description length of K-means as a compressor. Data $\mathbf{X}$ is a $d \times n$ matrix and cluster $\mathbf{C}$ is a $d \times k$ matrix where d is the number of data dimensions, n is the number of data points and k is number of clusters:

$$\begin{aligned} \mathbf{X}&: d \times n \text{ matrix} \\ \mathbf{C}&: d \times k \text{ matrix} \end{aligned} \qquad (1)$$

Here, we use the basic idea of minimum description length. The data is compressed by the dictionary vectors and residual vectors, and the description length is written as follows:

$$\begin{aligned} &(\text{"description length" of dataset}) = \\ &(\text{"data length" of dictionary vectors}) + \qquad (2) \\ &(\text{"data length" of residual vectors}) \end{aligned}$$

In this paper, we present two types of definition to represent the length of vectors. In the simpler formulation, the sum of square of individual elements of $\mathbf{X}$ is used as the data length (DL1). In the other formulation, the number of digit required to describe the square sum (in base 2 logarithm) is used as the data length (DL2):

$$(\text{DL1 of } \mathbf{A}) = |\mathbf{A}|_2^2 \tag{3}$$

or

$$(\text{DL2 of } \mathbf{A}) = n \log\left(\frac{|\mathbf{A}|_2^2}{nh} + 1\right) \tag{4}$$

where $\mathbf{A}$ is $d \times n$ matrix and $h$ is an arbitrary number of quantization unit and

$$|\mathbf{A}|_2^2 = \sum_{i,j} A_{ij}^2. \tag{5}$$

To prevent divergence, 1 is added in the parentheses of eq. (4).

Using the above two definitions of description length (DL1 and DL2), we define compression ratios of k-means (hereafter, we call these KMCR1 and KMCR2). We define KMCR1 as follows:

$$(\text{KMCR1}) = \frac{|\mathbf{C}|_2^2 + |\mathbf{R}|_2^2}{|\mathbf{X}|_2^2} \rightarrow \left(\frac{k}{n}|\mathbf{X}|_2^2 + |\mathbf{R}|_2^2\right)/|\mathbf{X}|_2^2, \tag{6}$$

where $\mathbf{R}$ is the matrix of residual vectors after clustering. The arrow in eq. 6 indicates that $\frac{k}{n}|X|_2^2$ is used for evaluating the length of $\mathbf{C}$ instead of $|\mathbf{C}|_2^2$. This is to prevent the DL1 of cluster center vectors becomes too small when the cluster center vector is near the origin. We assume that the cluster center vectors $\mathbf{C}$ are contained in the prepared memory whose size is defined by original data vectors X. Note that the value of KMCR1 is equal to 1 for k=0 and k=n.

Next, KMCR2 is defined using DL2 as below:

(KMCR2)
$$= \frac{k \log\left(\frac{|\mathbf{X}|_2^2}{nh} + 1\right) + n \log\left(\frac{|\mathbf{R}|_2^2}{nh} + 1\right) + 2n \log k}{n \log\left(\frac{|\mathbf{X}|_2^2}{nh} + 1\right)} \qquad (7)$$

The basic idea of KMCR2 is that the digit number for representing the value of each component of matrices is used to evaluate the memory sizes which may be used to contain cluster and residual vectors. The last term in the numerator of eq. (7) is the memory size for labeling and we can ignore the term when the number of clusters k is sufficiently small compared to the data dimension d. When the number of clusters is large enough, we do not ignore the description length of the indices.

Our main proposal is that these criteria can be used as a systematic method to determine the number of clusters for the k-means algorithm. We explore how these criteria behave in experiments with synthetic and real data in practice. As we show later in the section of numerical examples the behavior of the criteria strongly depends on the nature of data (e.g. dimensions, number of data and its distribution). In other words, which criterion matches best for a given dataset is a priori unknown as much as the true clustering and probabilistic distribution is unknown for a given dataset. As we show in the section of numerical examples, KMCR2 tends to yield a smaller number of clusters, whereas KMCR1 shows a smaller number of clusters for the simple 2D clustering example.

2.3. Review of K-means algorithm

The algorithm of k-means is reviewed as below:

(1) Set initial cluster $\mathbf{C}$ (d × k matrix), and $\mathbf{c}^{(j)}$ is vector component of $\mathbf{C}$.

(2) Find appropriate cluster index j for each $\mathbf{x}^{(i)}$ ($\mathbf{x}^{(i)}$ is vector component of $\mathbf{X}$), s.t. $\left|\mathbf{c}^{(j)} - \mathbf{x}^{(i)}\right|_2^2$ is minimum

(3) Revise new cluster $\mathbf{C}$ using matrix the $n \times k$ matrix $\mathbf{E}$ defined by:

$$E_{ij} = 1 \text{ (for the case that i − th data is a member of j − th cluster)}$$
$$= 0 \text{ (otherwise)},$$
$$\mathbf{C} \leftarrow \mathbf{XES}^{-1},$$

where $\mathbf{S}$ is $k \times k$ diagonal matrix whose diagonal elements are summation of $\mathbf{E}$ (e.g. $\mathbf{S} = \text{diag}(\text{sum}(\mathbf{E}))$; sum(E) is the Matlab expression for summation along the first array dimension, and diag is the Matlab expression for constructing a diagonal matrix from a vector). Some of the diagonal elements $\mathbf{S}$ become zero after iteration, for such case, the element is excluded ($k$ is reduced while the iterations).

(4) Return to (2) until convergence.

(5) After convergence, to compute the above compression ratios and 2nd stage criteria (shown below), calculate matrix of residual vectors, $\mathbf{R} = \mathbf{X} - \mathbf{CE}$.

## 2.4. Redundancy check

K-means sometimes tends to make more clusters than the actual number of clusters of the dataset (even for clean synthetic dataset) when the initial value of k is chosen too large. For such a case, there may be many redundant cluster vectors, so the set of cluster center vectors may also be compressed by k-means. To check the redundancy, we use a 2-stage compression procedure. If the cluster center vectors can be compressed, such cluster center vectors have redundancy and the number of clusters is not desirable. In other words, to prevent redundancy in the cluster vectors the compression ratio of cluster vectors matrix $\mathbf{C}$ should be roughly equal to 1. Assume that $\mathbf{R_1}$ is the residual of 1st stage compression (note that $\mathbf{R_1} = \mathbf{R}$ in eqs. 6 and 7) and $\mathbf{R_2}$ is the residual of 2nd stage compression, and $k_2$ is the number of clusters for the cluster center vectors. We can calculate the description length of the second stage as:

$$\text{(2nd stage of KMCR1)} = \left(\frac{k_2}{n}|\mathbf{X}|_2^2 + |\mathbf{R_2}|_2^2 + |\mathbf{R_1}|_2^2\right)/|\mathbf{X}|_2^2 \tag{8}$$

and

(2nd stage of KMCR2)

$$= \left(k_2 \log\left(\frac{|\mathbf{X}|_2^2}{nh} + 1\right) + k \log\left(\frac{|\mathbf{R}_2|_2^2}{kh} + 1\right) \right.$$
$$\left. + n\log\left(\frac{|\mathbf{R}_1|_2^2}{nh} + 1\right)\right) / n \log\left(\frac{|\mathbf{X}|_2^2}{nh} + 1\right). \qquad (9)$$

Plotting both 1st stage KMCR1/2 and 2nd stage KMCR1/2, we can determine whether the 1st stage clustering is redundant or not.

2.5 Procedure

The procedure we adopt in this paper is as follows:

1. Run k-means for various k, and calculate KMCR1 and KMCR2 the 2nd stage k-means, and the compression ratios. Find the cluster numbers that minimize KMCR1 and KMCR2. Output the numbers and cluster index.

2. If necessary, prepare dense set of k and re-run k-means.

Note that each process of K-means is operated very fast by using "litekmeans" (Michael Chen, provided by Matlab central), which is about 50 times as fast as the Matlab native k-means algorithm. We create a wrapper for "litekmeans" to compute the criteria for various initial clusters. Note that the above procedure can also be used for other criteria of description length.

2.6 Data

We used 511 T1-weighted (T1w) structural images from the Human Connectome Project (HCP) (e.g. Glasser et al. 2013). In this study we preprocessed T1w images through a standard preprocessing procedure for Voxel Based Morphometry (VBM) implemented in Statistical Parametric Mapping (SPM: Friston et al, 2007). The T1w images were first segmented into grey matter, white matter, and cerebrospinal fluid using SPM8 (http://www.fil.ion. ucl.ac.uk/spm). A Diffeomorphic Anatomical Registration Through Exponentiated Lie Algebra (DARTEL) was performed for registering segmented grey matter images across individuals [(Ashburner & Friston, 2000; Ashburner, 2007)](). The co-registered images were further transformed to the standard MNI space while voxel intensities were modulated by the Jacobian determinants of the deformation fields computed by DARTEL. The registered images were smoothed with a Gaussian kernel (FWHM = 8 mm) and were then transformed to MNI space. This preprocessing ensured that the intensity of each voxel in the normalized space corresponds to the local grey matter volume. We further normalized the grey matter volume by the sum of total grey matter and white matter volumes. The resulting grey matter volume images had about 400,000 degrees of freedom (or voxels per image), and in this paper, we filtered out voxels outside the brain using a threshold (<0.1) and 3D data were downsampled by a factor of 5 for each of the xyz directions to reduce the dimension. We arranged each grey matter voxel data in the following matrix form:

$$A: N_p \times d \text{ matrix,}$$

where $N_p$ is the number of participants, and d is the number of downsampled voxels (d=3416, about 15000). Hereafter, we refer to the brain data downsampled by a factor of 5 and 3 as A5 and A3, respectively. We normalized the data per voxel by computing the z-score of A per column and took the following inner product to compute the covariance matrix:

$$X = Z^T Z \ : d \times d \text{ matrix}.$$

We used X to make a brain parcellation model later.

3. Numerical examples

In the following numerical examples, we used a single workstation (dual Xeon E2600, 128GB memory).

3.1 Behavior of the criteria

To show how these criteria work, numerical experiments for simple cases were performed. We create a synthetic dataset using parameters, n: number of data points, $d$: dimension of data, $k_c$: number of clusters and $n_c$: number of members in each cluster, respectively. Cluster centroids are located on the d-dimensional sphere (radius=1), and $n_c$ points are randomly distributed inside the sphere (e.g. radius=0.01). We use $n_c = 50$ and $k_c = 300$, this results in a total of 15000 data points. We choose d=2, 200, and 10000 for numerical experiments.

The computational results of KMCR1 and KMCR2 are shown in Figure 1. The codes for the KMCR computation and the experiments are available from https://github.com/mizutaniaraya/computeKMCR). The results for the synthetic datasets with d=2, 200, and 10000 are shown in Figure 1a, 1b and 1c respectively. Figures 1d and 1e show the results for the HCP brain data downsampled by a factor of 5 and 3, respectively We can see that in all cases KMCR1 and KMCR2 have minima within the tested range of cluster numbers. For the case d=10000 (figure 1c), the minima of KMCR1 and KMCR2 both indicate $k_c > 300$. This is because especially when data dimensionality is high k-means tends to find redundant cluster vectors compared to those expected from the true distribution. In each plot, red points indicate the criteria of 2nd-stage compression for various numbers of clusters. For the cases where the number of clusters was small, the compression ratio did not change by the 2nd-stage compression, but for a greater number of clusters, the values become smaller than those of 1st stage compressions reflecting the tendency of k-means to include redundant cluster vectors.

In Figure 2, we show the behavior of KMCR1 and KMCR2 in a single diagram (d=2, 200 and 10000 of synthetic dataset are plotted by blue, red and green lines, respectively, and the A5 and A3 cases of HCP brain data by grey and black lines). For synthetic datasets, as k increased, the KMCR1 and KMCR2 decreased and turned to increase. For the synthetic datasets in this diagram, KMCR1 first reached the minimum value, before KMCR2 reached its minimum. For the HCP brain image datasets, the orientation of the trajectory was different: KMCR2 reached the minimum before KMCR1. These experiments with synthetic and biological data suggest that the shape of the trajectories depend on the characteristics of the dataset such as the dimensionality, sample size, the shape of the distributions of cluster members, the magnitude of random noise components and so on. While the combination of these factors collectively contributes to the behavior of KMCR1 and KMCR2, the diagram presented here may prove useful for capturing the differences of the characteristics of dataset.

In Figure 3, we show the dependence of KMCR2 on the quantization factor h (1/100, 1, 100) for the synthetic dataset with d=200 and the A3 case of HCP. For both cases, as h increases, the values of KMCR2 become smaller but the shapes of the trajectories seem to be almost unchanged regardless to the value of h.

3.2 Results of brain parcellation using HCP data

We performed a parcellation of brain regions using k-means clustering on HCP brain image datasets. In figure 4, the result of parcellations for the dataset downsampled by 5 and 3 (A5 and A3) are visualized. The result for A5 is mostly symmetric, whereas the result for A3 showed more asymmetry in comparison. The slight asymmetry revealed in our analysis is consistent with a previous report (e.g. van Essen et al., 2012). The numbers of clusters which minimized KMCR1 were 201 (A5) and 473 (A3), and the numbers of clusters which minimized KMR2 were 20 (A5) and 171 (A3). The number of clusters was larger for A3 than that of A5, because A3 has more detailed information than A5.

In figure 5, we show the initial value dependence of brain clustering using A3 data. We used a cluster number of 100, and tested three initial cluster conditions: randomly scattered near the center of the whole brain (figure 5a), randomly scattered centered within each hemisphere (figure 5b) and randomly scattered around four locations (left-right x anterior-posterior) (figure 5c). For each experiment, even if the initial cluster centroids were located at the center of the brain, the centroids were distributed in the YZ plane immediately at the first iteration. The centroids remained around the midline even after the final iteration. This reflects the fact that most clusters determined by our analysis were symmetrical across the hemispheres (see Figure 4).

We executed k-means clustering by 100 different initial clusters and projected the centroids on the YZ plane (Figure 6). The centroids for different initial values are plotted by different colors. The exact positions of the centroids were different for each experiment, but the final results showed a clear lineation of centroids. From this, it can be surmised that the k-means clustering depends on the initial clusters but the resulting clusters are found within restricted space for the brain data. The implication is that the clusters obtained by k-means clustering efficiently reduce the dimensionalities of the original brain image dataset, even if it has dependence on the initial values. We conclude that the k-means clustering is usable for the dimensionality reduction to make some prediction models of human personality traits using brain image dataset.

4. Conclusion

In this paper, we proposed that data compression ratios measured by description length be used as criteria for choosing an appropriate number of clusters in k-means algorithms. Specifically, we proposed two types of criteria, KMCR1 and KMCR2 based on two types of description length measures. Our numerical experiments both with synthetic data and empirical data from neuroimaging suggest that this approach could be used for practical purposes, though the estimated number of clusters depended various factors such as the dimensionality of data, and which criterion was used. Our experiments with synthetic data suggest that the true cluster number can deviate from that suggested by our criteria. This is caused by strong dependency of our criteria on the characteristics of dataset, which is generally difficult to estimate from observed data. Notwithstanding these limitations, our proposed criteria can offer a useful guidance to determine the number of clusters in a principled manner. In particular, our experiment with MRI dataset showed that the criteria are usable for a brain parcellation problem. While the k-means clustering depended on the initial values, the dependence turned out to be limited for the brain parcellation problem. In future work, this scheme might be used for dimensionality reduction to boost the accuracy of prediction models of an individual's traits from brain MRI data.

Acknowledgements

We thank Martin Biehl for reading the manuscript and giving many critical comments.

References

Ashburner, J. (2007). A fast diffeomorphic image registration algorithm, Neuroimage, 38, 95—113 .

Ashburner, J. and Friston,K.J., (2000). Voxel-based morphometry – the methods, Neuroimage,11 805—821.

Friston, K.J., Ashburner, J., Kiebel, S.J., Nicholus, T.E. and Penny, W.D., (2007). Statistical Parametric Mapping: The Analysis of Functional Brain Images. Academic Press.

Glasser, M.F., Sotiropoulos, S.N., Wilson, J.A., Coalson, T.S., Fischl, B., Andersson, J.L., ⋯ WU-Minn HCP consortium, (2013). The minimal preprocessing pipelines for the human connectome project, Neuroimage, 80, 80—104.

Glasser, M.F., Coalson T.S., Robinson, E.C., Hacker, C.D., Harwell, J., Yacoub, E., ⋯, van Essen, D.C., (2016). A multi-modal parcellation of human cerebral cortex, Nature, 536, 171—178. DOI:10.1038/nature18933.

Kanai, R., and Rees, G. (2011). The structural basis of inter-individual differences in human behaviour and cognition. Nature Reviews Neuroscience 12 (4), 231-242.

Pelleg, D, and Moore A., (2000). X-means: Extending K-means with Efficient Estimation of the Number of Clusters. ICML.

Sun, W., Wang, J. and Fang, Y., (2012). Regularized k-means clustering of high-dimensional data and its asymptotic consistency, Electronic Journal of Statistics, 6, 148–167, DOI: 10.1214/12-EJS668.

van Essen, D.C., Glasser, M.F., Dierker, D.L., Harwell, J. and Coalson, T., (2012). Parcellation and hemispheric asymmetries of human cerebral cortex analyzed on surface-based atlases, Cereb. cortex 22, 2241—2262.

(a) synthetic d=2

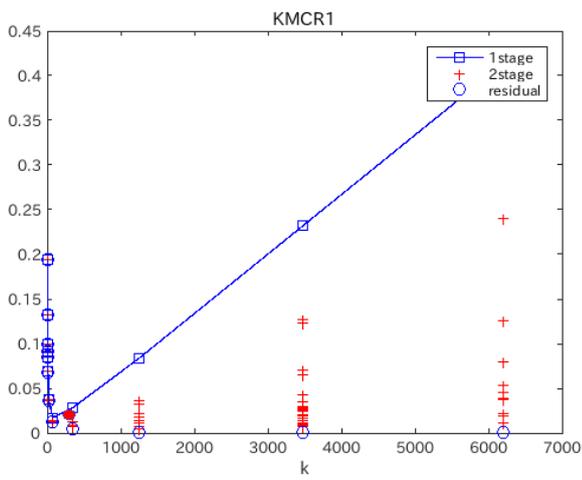
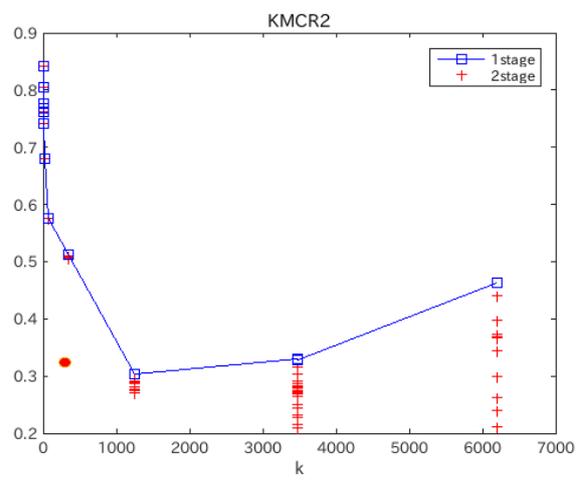

(b) synthetic d=200

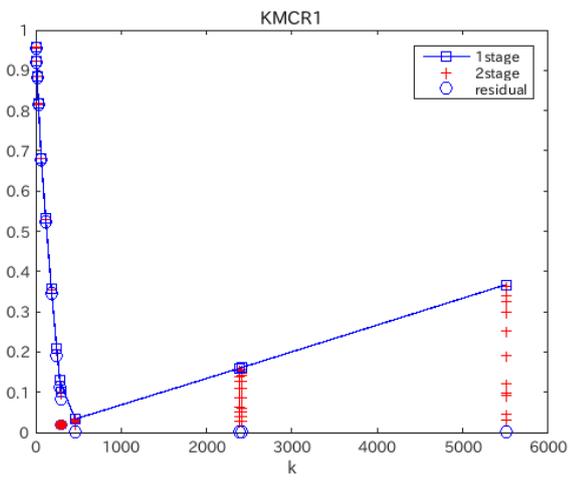
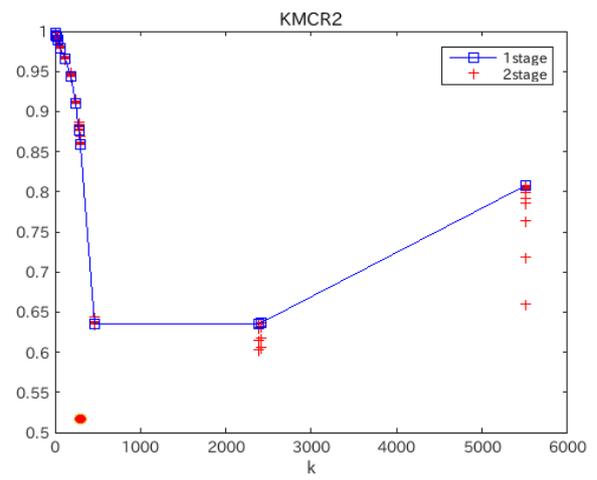

(c) synthetic d=10000

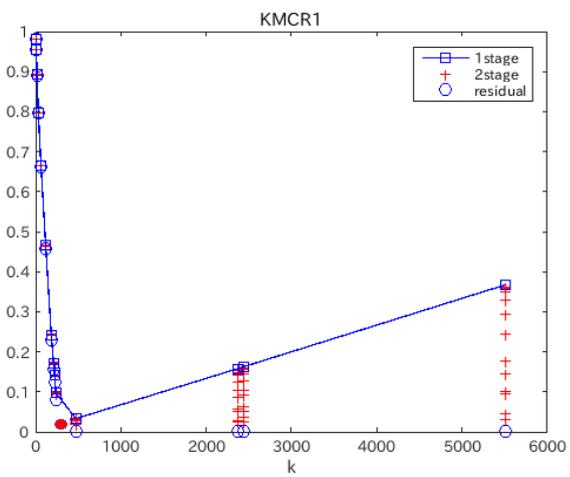
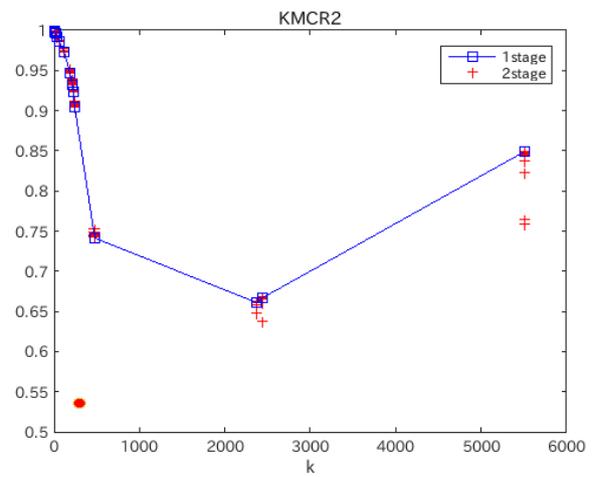

(d)HCP brain downsampled by a factor 5

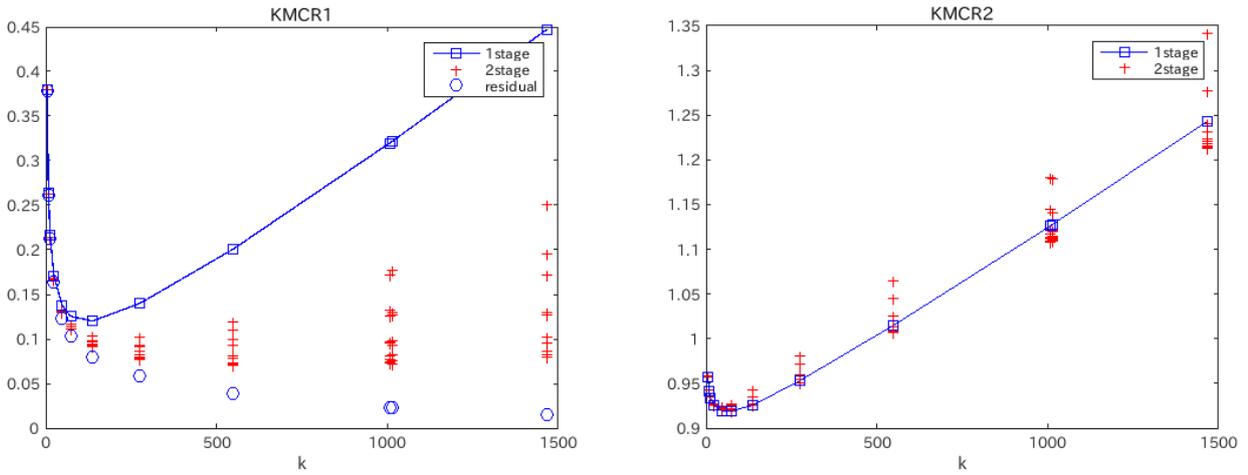

(e)HCP brain downsampled by a factor of 3

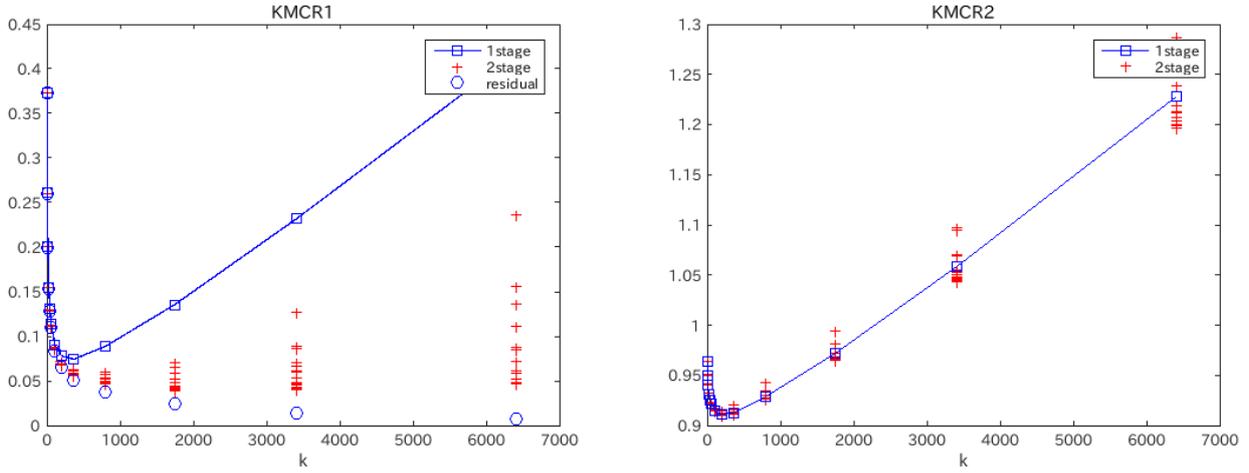

Figure 1

Compression ratio criteria KMCR1 (left) and KMCR2(right) for synthetic dataset (a-c) and HCP brain image dataset (d and e) are shown. The horizontal axis shows the number of cluster used (k) and the vertical axis shows the compression ratio. Blue squares and line show the values computed by the 1-stage algorithm and red crosses show the values computed by the 2-stage algorithm. Blue circles show the residual normalized by square sum of the dataset. For the synthetic dataset, the value of true clustering are plotted by red filled circle at k=300. For KMCR2, h=1 is used. For a large value of k, the 2nd stage algorithm could further compress the vectors indicating that the cluster vectors still had some redundancy.

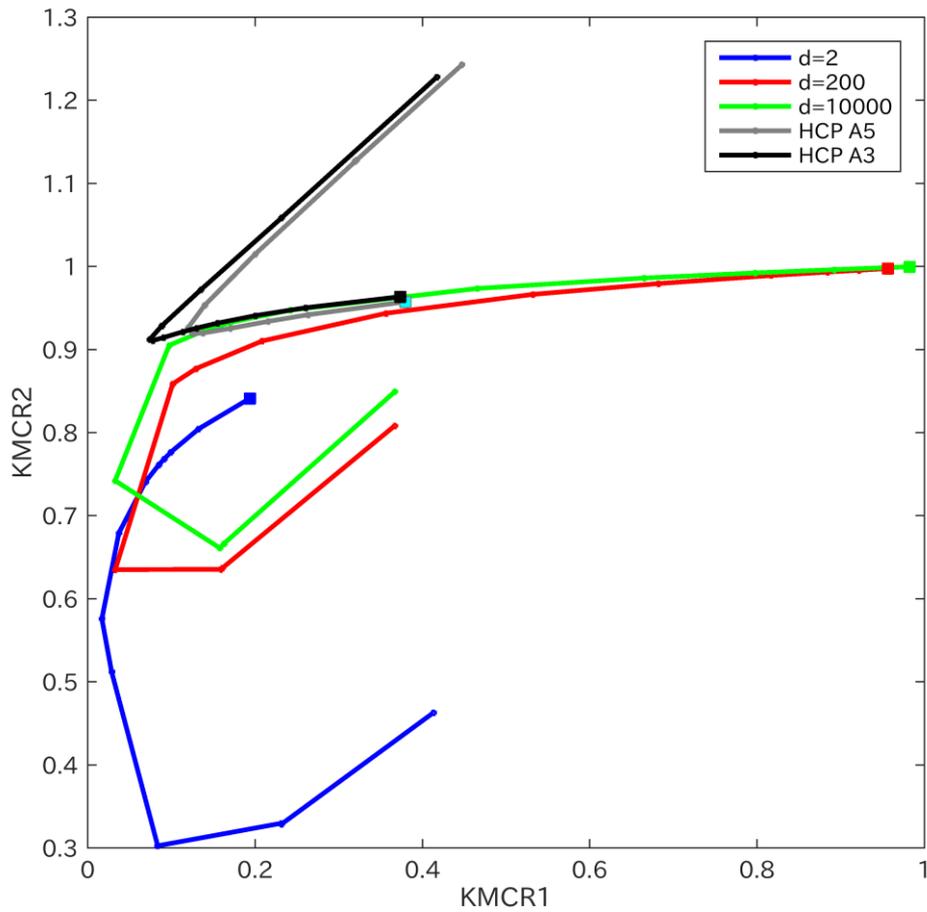

Figure 2

KMCR trajectories for synthetic and HCP brain image datasets. The horizontal axis shows KMCR1 and the vertical axis shows KMCR2. The trajectories for the synthetic datasets are plotted in blue (d=2), red (d=200) and green (d=10000) lines, respectively, and the HCP brain image datasets are also plotted in   grey (A5) and black (A3). Filled square marks indicate the points for k=4.

The shape of trajectories depends on the nature of datasets. The trajectories of synthetic datasets were similar to each other in the overall shape, but are distinct from those for the brain datasets.

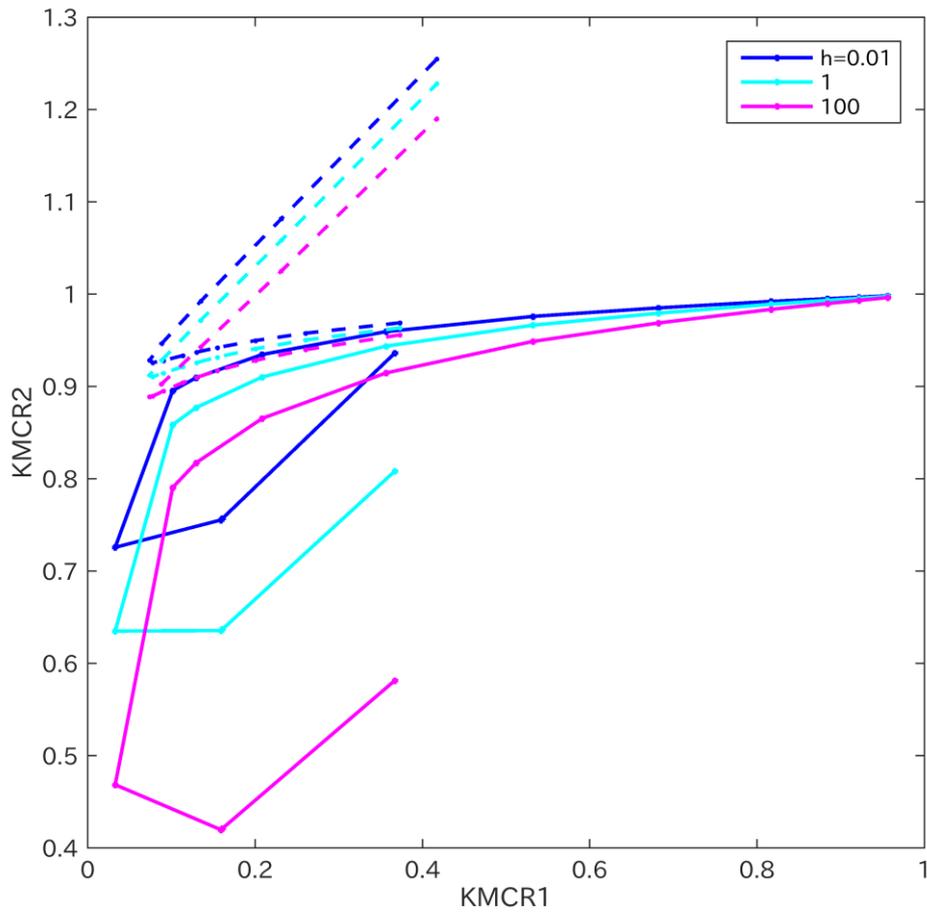

Figure 3

The dependence of compression ratios on quantization values h. Synthetic dataset for d=200 (solid line) and HCP brain (A3; broken line) cases are plotted for three different values of h (blue: 0.01, cyan:1, magenta:100). The shape of trajectories was similar among each other regardless of the values of h. The distinct overall differences in the shape of the trajectory between the synthetic and HCP datasets can be also seen in this graph (see Figure 2).

Figure 4a

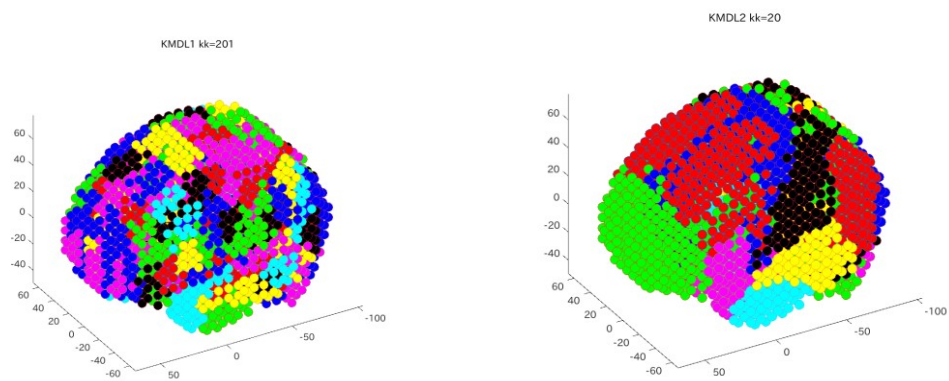

Figure 4b

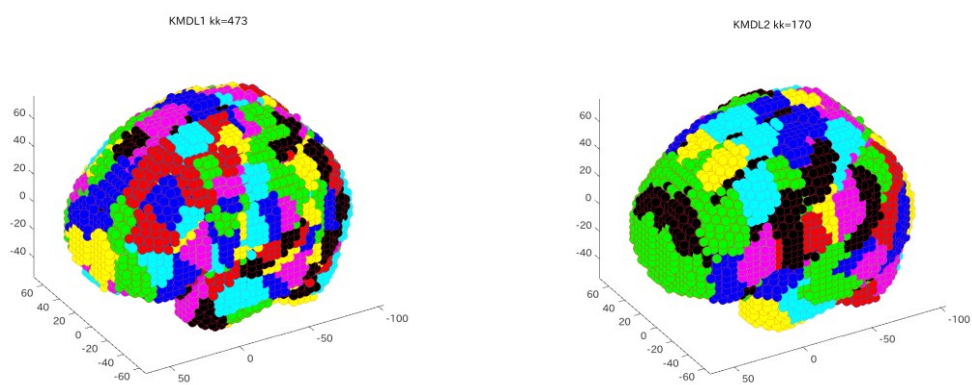

Fig. 4 Brain parcellation

(a) The result of clustering using dataset downsampled by 5 (A5), and (b) using dataset downsampled by 3 (A3). The left panel is the model that minimized KMCR1 and the right is the model that minimized KMCR2.

Figure 5a

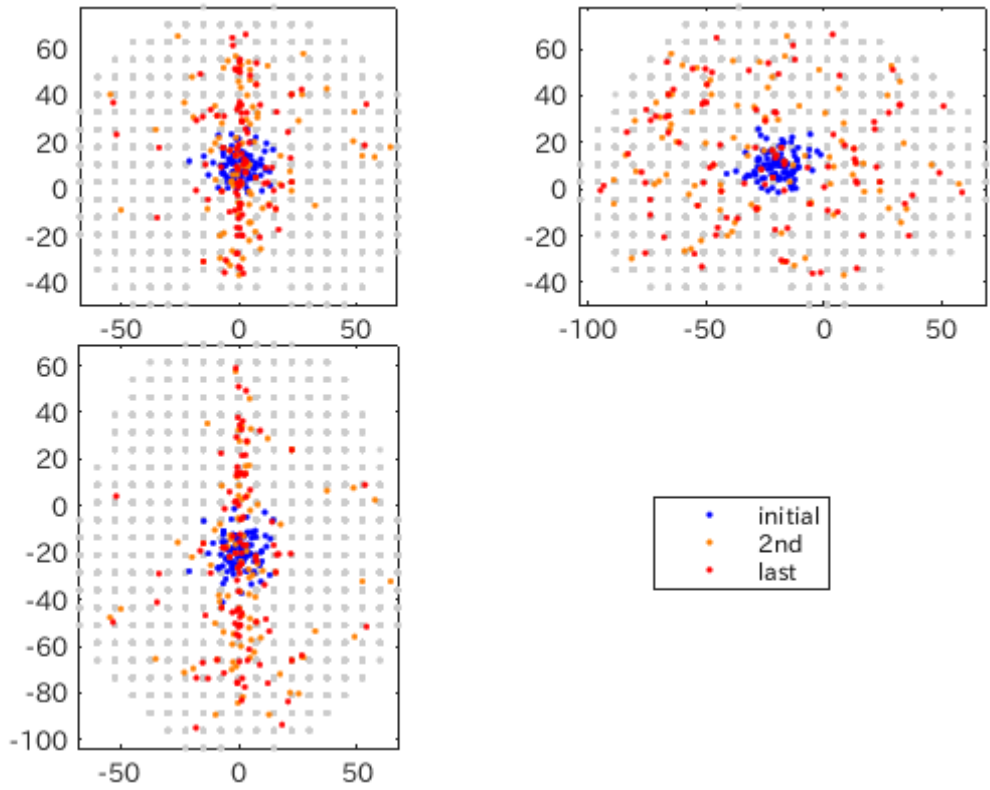

Figure 5b

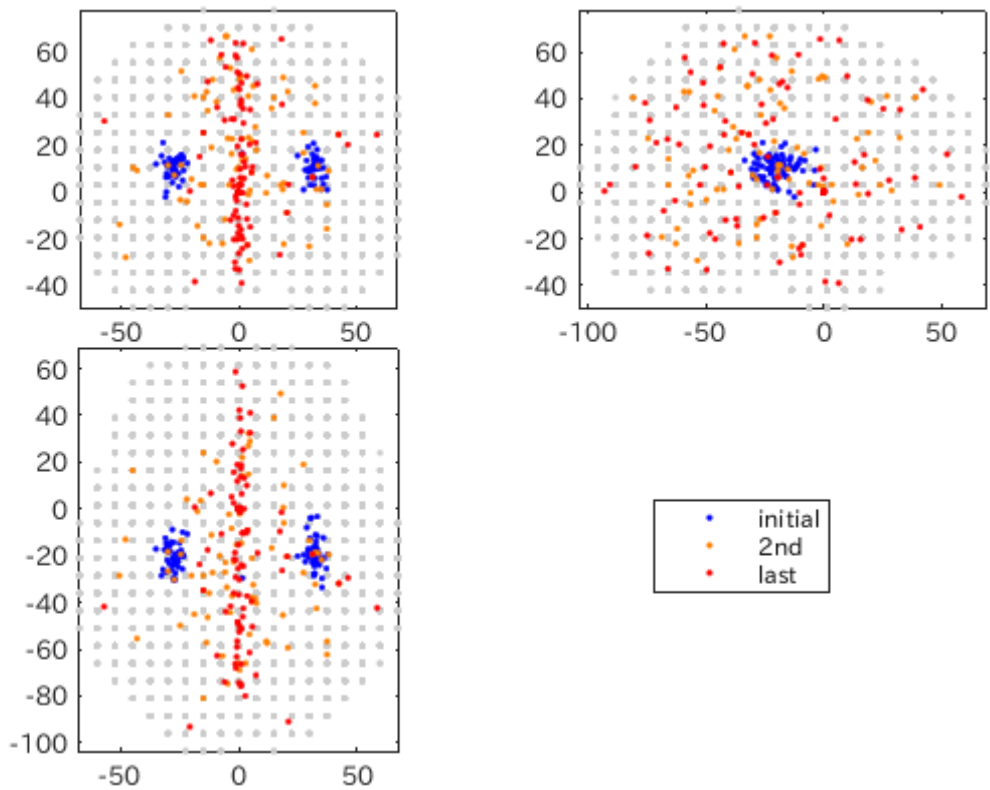

Figure 5c

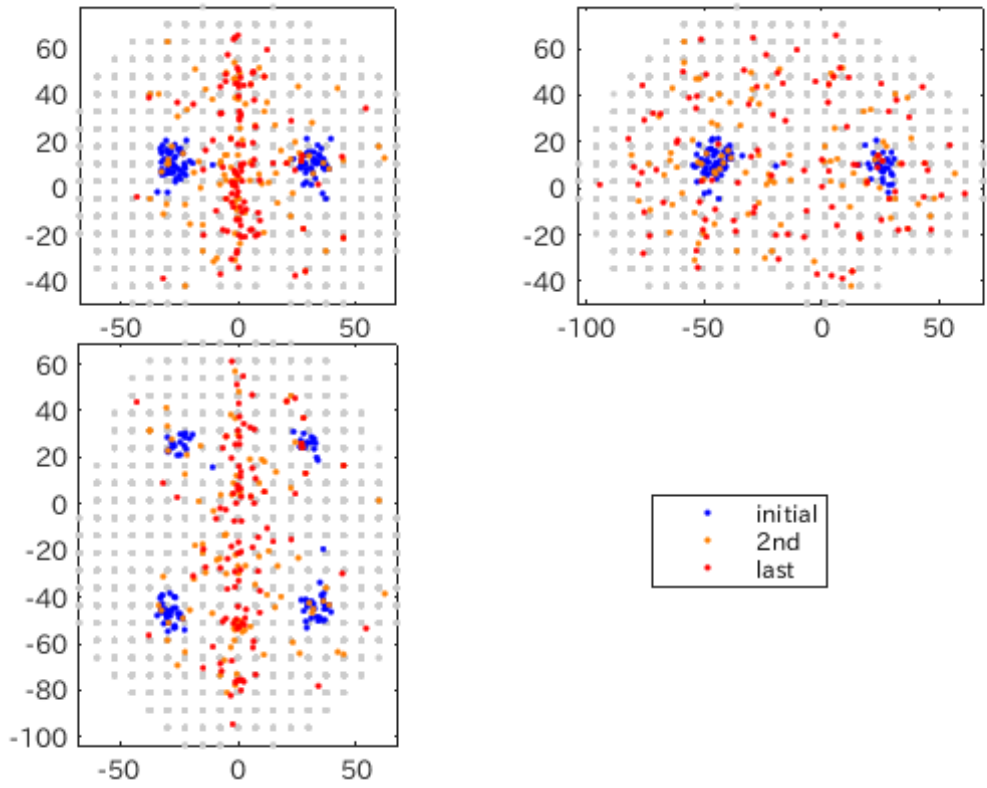

Figure 5

Dependence on initial clusters. The coronal, horizontal and saggital (top-left, bottom-left, top-right respectively) projection is shown. The initial cluster centroids are plotted by blue, the cluster centroids at 2nd iteration are plotted by orange, and the cluster centroids at the last are plotted by red. The panels (a—c) are the results which started from different initial value distributions.

Figure 6

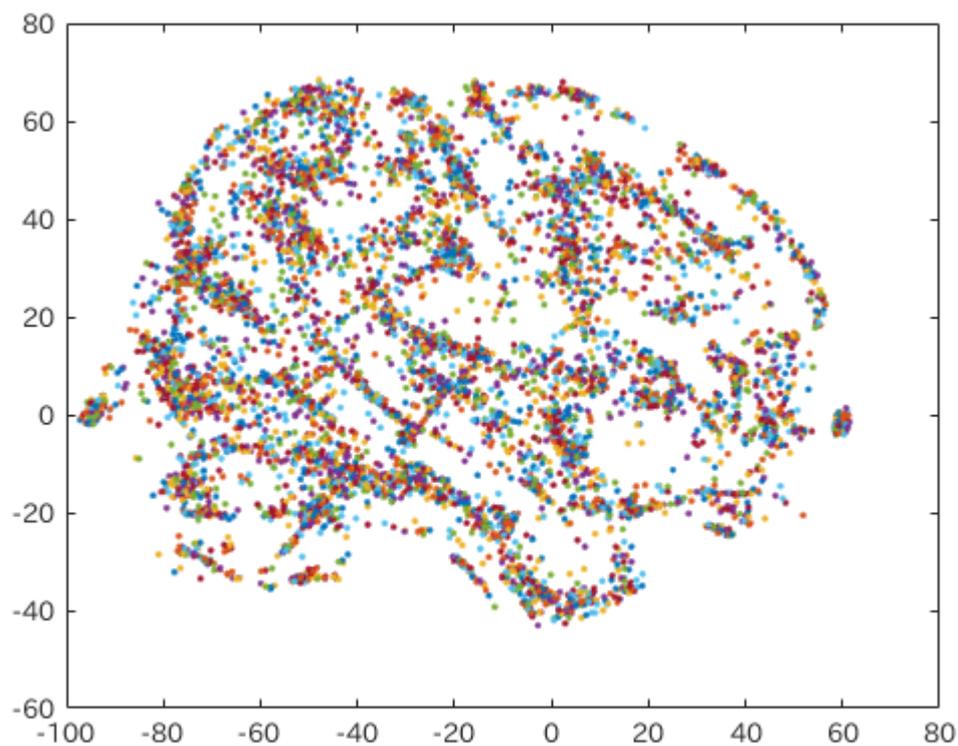

Figure 6

Cluster centroids of 100 different runs projected on YZ plane. Cluster number k=100 is used for this experiment. The different colors of points indicate different runs. We can see clear lineation of cluster centroids. The cluster centroids are gathered into reduced dimensionalities regardless of the initial clusters.